\documentclass[sigconf]{acmart}

\usepackage{pifont}
\usepackage{pbox}
\usepackage{enumitem}
\usepackage{multirow}
\usepackage{amsmath,amsfonts}
\usepackage{dsfont}

\AtBeginDocument{%
  \providecommand\BibTeX{{%
    \normalfont B\kern-0.5em{\scshape i\kern-0.25em b}\kern-0.8em\TeX}}}

\setcopyright{acmcopyright}

\copyrightyear{2023}
\acmYear{2023}
\setcopyright{rightsretained}
\acmConference[MiLeTS '23]{9th ACM SIGKDD International Workshop on Mining and Learning From Time Series}{August 2023}{Long Beach, CA, USA}
\acmBooktitle{9th ACM SIGKDD International Workshop on Mining and Learning From Time Series, August 2023, Long Beach, CA, USA}
\acmDOI{}
\acmISBN{}

\makeatletter
\gdef\@copyrightpermission{
    \begin{minipage}{0.3\columnwidth}
    
    \href{https://creativecommons.org/licenses/by-sa/4.0/}{\includegraphics[width=0.90\textwidth]{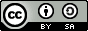}}

    \end{minipage}\hfill
    \begin{minipage}{0.7\columnwidth}
    
    \href{https://creativecommons.org/licenses/by-sa/4.0/}{This work is licensed under a Creative Commons Attribution-ShareAlike International 4.0 License.}
    
    \end{minipage}
    \vspace{5pt}
}
\makeatother

\begin{document}


\title{Learning Behavioral Representations of Routines From Large-scale Unlabeled Wearable Time-series Data Streams using Hawkes Point Process}


\author{Tiantian Feng}
\affiliation{%
\institution{University of Southern California}
\city{Los Angeles}
\state{CA}
\country{USA}
}
\email{tiantianf@usc.edu}

\author{Brandon Booth}
\affiliation{%
\institution{University of Colorado Boulder}
\city{Boulder}
\state{CO}
\country{USA}
}
\email{brandon.booth@colorado.edu}

\author{Shrikanth Narayanan}
\affiliation{%
  \institution{University of Southern California}
  \city{Los Angeles}
  \state{CA}
  \country{USA}  
}
  \email{shri@sipi.usc.edu}

\renewcommand{\shortauthors}{Tiantian Feng et al.}
\renewcommand{\shorttitle}{Learning Behavioral Representations of Routines \\From Large-scale Unlabeled Wearable Time-series Data Streams using Hawkes Point Process}

\begin{abstract}
Continuously-worn wearable sensors enable researchers to collect copious amounts of rich bio-behavioral time series recordings of real-life activities of daily living, offering unprecedented opportunities to infer novel human behavior patterns during daily routines. Existing approaches to routine discovery through bio-behavioral data rely either on pre-defined notions of activities or use additional non-behavioral measurements as contexts, such as GPS location or localization within the home, presenting risks to user privacy. In this work, we propose a novel wearable time-series mining framework, Hawkes point process On Time series clusters for ROutine Discovery (HOT-ROD), for uncovering behavioral routines from completely unlabeled wearable recordings. We utilize a covariance-based method to generate time-series clusters and discover routines via the Hawkes point process learning algorithm. We empirically validate our approach for extracting routine behaviors using a completely unlabeled time-series collected continuously from over 100 individuals both in and outside of the workplace during a period of ten weeks. Furthermore, we demonstrate this approach intuitively captures daily transitional relationships between physical activity states without using prior knowledge. We also show that the learned behavioral patterns can assist in illuminating an individual's personality and affect.
\end{abstract}





\maketitle

\section{Introduction}
\label{sec:introduction}

Wearable sensors have garnered considerable interest in many fields, such as healthcare, user authentication, and entertainment, over the last two decades \cite{patel_sensor_review, booth2019multimodal, saeb2015mobile}. These non-obtrusive devices, which are often small in size and have efficient computational capabilities, can be extremely useful in capturing vital bio-metric and bio-behavioral data from individuals over a prolonged period in natural settings \cite{stress_physio_sano, avramidis2022multimodal}. Such rich and vast amounts of multimodal time-series data collected directly from everyday life allow for a more comprehensive understanding of the factors affecting activities of daily living (ADLs) \cite{adl_lawton}, including but not limited to social interactions \cite{wear_social_mohammad, jati2021temporal}, sleep patterns \cite{sleep_roebuck, feng2021multimodal}, physical activities \cite{feng2021multimodal, booth2019toward}, and even emotion variations \cite{emotion_hui}. Increasingly, the ability to recognize ADLs offers researchers opportunities to investigate broad human behavior patterns and infer common daily routines. Routine behavior is notably meaningful in quantifying what activity pattern people adopt and whether these patterns cause variations of psychological well-being and personality within groups of people \cite{routine_baldwin}.

In this paper, we present a novel data processing approach, \textbf{H}awkes point process \textbf{O}n \textbf{T}ime series clusters for \textbf{RO}utine \textbf{D}iscovery (\textbf{HOT-ROD}) for learning routine patterns in biobehavioral time series from wearable sensors. Our proposed HOT-ROD pipeline includes data processing components ranging from aggregation, imputation, filtering, time-series clustering, and routine discovery. We show that the proposed routine features, comprised of temporally linked cluster transitions in multimodal wearable recordings, can assist in illuminating an individual's personality and affect as well as aspects of task performance such as job behaviors. The main contributions of this work are as follows:

\begin{itemize}[leftmargin=*]
    \item We propose a novel combination of Toeplitz Inverse Covariance-Based Clustering (TICC) \cite{hallac2017toeplitz} and Hawkes point process \cite{hawkes_xu} for discovering routine characteristics in long-term real-world wearable recordings. The approach can operate without expert knowledge of the data or the collection of sensitive contextual information.

    \item Our learned routine patterns capture temporal relationships between adjacent time-series clusters, thus providing valuable insights into understanding the shared behavior patterns within a group of individuals.

    \item Using naturalistic heart rate and step count features from over 100 individuals in the workplace and at home over a period of ten weeks, we show that our HOT-ROD approach combined with daily summaries of physical activity from wearable sensors helps identify job properties and personalities of participants. Furthermore, we empirically show that our approach can achieve modest improvement in predicting individual attributes from a few days of recordings.

\end{itemize}

\section{Related Works}
\label{label:related_work}

Many conventional approaches in characterizing daily routines require the acquisition of labeled contexts, like trajectory in home settings \cite{trajectory_endo}, life event sequences (eating, sleeping, etc.) \cite{life_seq_dawadi}, and GPS locations \cite{gps_michael}. The Kasteren data set \cite{home_setting} was collected in a $3$-bedroom apartment setting for a period of $28$ days using $14$ state change sensors. The investigators then achieved a timeslice accuracy of $95.6\%$ on this data set using a hidden Markov model and conditional random fields. Some following studies have successfully utilized a probabilistic neural network learning model to separate the normal routine from unusual and suspected routines \cite{home_setting_neural}. Meanwhile, other researchers have studied human behavior by tracking the spatial properties of participants via GPS \cite{gps_michael}. However, one primary concern about these studies is that the data collection protocol could invade privacy by tracking sensitive and identifiable knowledge about an individual, such as continuous GPS. These approaches might also be costly and not scalable due to the substantial amount of effort required from researchers in either setting up the recording system or annotating the data.

To prevent the acquisition of personally identifiable information, there have been a number of studies in establishing machine learning models to infer a pre-selected set of activities (walk, stand, etc.) from unlabeled wearable time-series \cite{activity_wadley}, like motion and posture. However, these models are typically trained from data gathered in laboratory settings and may not yield good performances on in situ data sets. As an alternative, motif-based methods have obtained empirical success in detecting repeated patterns; but this method can be computationally prohibitive due to the optimal granularity for motif patterns that must be searched through the whole time-series. Unlike the motif-based data mining method, one other recent study proposed to learn routine behaviors via a sparse and low-rank matrix decomposition technique \cite{routine_yürüten}. In this work, the real-world physical activity data collected from Fitbit were used to cluster the behaviors of participants without expert knowledge or micro-pattern extraction. The main disadvantage associated with this approach is however the limited interpretability of decomposed matrices returned from sensor matrix operations. To our knowledge, our proposed causality-based pattern extraction from unlabeled wearable sensor recordings has not been previously considered.

\section{Dataset Introduction}
\label{sec:dataset}

In this study, we used a dataset called TILES-2018 \cite{mundnich2020tiles} that is publicly available for our experiment. This data set involves a set of comprehensive experiments with the target of studying how physiological and behavioral variables affect employee wellness, personality, and workplace stress. Throughout a ten-week period, physiological, environmental, and human interaction data were gathered from hospital employees who primarily provide patient care (nurses, technicians, etc.) at a large critical-care hospital. The complete dataset consisted of $213$ hospital workers comprised of $120$ female participants. In total, there were 113 (54.3\%) registered nurses enrolled in the study, with the rest reporting some other job title, such as occupational or lab technicians. A total of 54 participants reported working the night shift and the rest worked during the day. More details regarding the dataset can be referred to \cite{mundnich2020tiles}

\vspace{-2mm}
\subsection{Study Procedure}

The participants involved in this study need to complete a \texttt{Initial Assessment} survey during onboarding, consisting of a web-based series of surveys pulled from existing test battery questionnaires that assessed standard demographic information, personality, and affect variables. In this work, we primarily investigate how automatically discovered routine patterns correlate with job type, gender, shift type, personality, and affect. Each of the five personality factors (extraversion, agreeableness, conscientiousness, openness, neuroticism) is measured via the Big Five Inventory-2 survey \cite{big_five_soto}. Five-factor scores were computed by taking the average of all responses, where each factor score is in the range of $1$-$5$. The Positive and Negative Affect Schedule (PANAS) was administered \cite{panas_watson} to measure affect. The PANAS consists of $10$ positive affect items and $10$ negative affect items. Positive and negative affect scores were calculated by summing individual scores from each group (positive and negative) with higher scores representing higher levels of corresponding affect.

\subsection{Wearable Data}
In this study, researchers instructed participants to wear a Fitbit Charge $2$ \cite{fitbit}, an OMsignal garment-based sensor \cite{om_signal}, and a customized audio badge \cite{feng2018tiles} which collectively tracks heart rate, physical activity, speech characteristics, and many other human-centric signals. Participants were asked to wear the OMsignal garment and audio badge only during their work shifts due to the battery limitation of these devices. However, participants were instructed to wear a Fitbit sensor as often as possible throughout the $10$-week data collection period. In the present study, we focus on the Fitbit time series data since it is present for most participants both during and outside of their working hours. This data stream offers information about energy expenditure, sleep quality, step count and heart rate measured through photoplethysmography (PPG).

\section{Method}
\label{sec:method}

\begin{figure}[ht]
	\centering
	\includegraphics[width=\columnwidth]{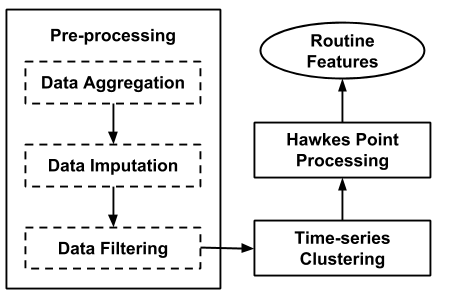}
    \caption{The data processing flow of Hawkes point process On Time-series for ROutine Discovery (HOT-ROD).}
    \vspace{-4mm}
    \label{fig:pipeline}
\end{figure}

In this section, we introduce our HOT-ROD analysis pipeline in discovering routine features from Fitbit time-series recordings. As shown in Fig.~\ref{fig:pipeline}, our proposed HOT-ROD data pipeline consists of three major modules: 1. \texttt{Pre-processing}; 2. \texttt{Time-series}; 3. \texttt{Hawkes point processing}. To calculate routine features using the Hawkes Point Process, we first group time-series by day. In this work, a day is defined as the variable period of time between sleep onsets as determined from the Fitbit daily summary. Sleep durations shorter than six hours (presumed to be naps) are ignored when determining these day boundaries in the time series. Some participants may sleep less than six hours regularly or may not wear their Fitbit devices to bed, which would result in measured days lasting upwards of 30 hours.  To remove these outliers, we maintain the data with its length of 20-28 hours according to the definition of a circadian cycle \cite{circadian_halberg}.

The \textit{pre-processing} component includes data aggregation, data imputation, and data filtering. We first aggregate multivariate time-series data at the fixed rate of one minute. The second step uses an Autoregressive integrated moving average (ARIMA) model to fill in the missing data in the aggregated output. We then utilize the Savitzky-Golay filter to smooth the imputed time-series data without substantially distorting the signal following previous work \cite{feng2019discovering}. Following the data pre-processing scheme, we cluster the pre-processed data stream using Toeplitz Inverse Covariance-Based Clustering (TICC) \cite{hallac2017toeplitz}. At the final stage, we extract routine features utilizing the Hawkes point process technique \cite{hawkes_xu} where cluster transitions serve as the process events. The details of each module are further described below.

\subsection{Data Pre-processing}

\noindent\textbf{Data Aggregation} Fitbit Charge $2$ sensors read PPG heart rate samples at intervals of less than one minute, but the time differences between two consecutive samples are inconsistent. Prior studies have suggested that PPG-based heart rate should be averaged over a one-minute duration to obtain a reliable measurement \cite{ppg_clinical}, and thus adopt this strategy. Another compelling reason to aggregate the PPG heart rate samples is to rate-match the data output of step count samples, which are also made available every minute.

\noindent\textbf{Data Imputation} Missing data in wearable sensors recordings are unfortunately unavoidable and are often encountered for various reasons including intermittent disconnections, body movements, and firmware malfunctions \cite{zhang_data_loss}. In this work, we select an Autoregressive integrated moving average (ARIMA) model to impute missing values \cite{arima}. We utilize ARIMA to populate missing values based on past observations. Missing points that occur in the first five data points in the time series are filled with mean values from the corresponding day in the time series. Prior literature \cite{imputation_dong, feng2019imputing, che2018recurrent} suggests that imputation works better when the proportion of missing data is small, thus we experimentally choose to fill the missing segments that are not continuously missing over 15 minutes. In this imputation experiment, we masked 10\%, 25\%, and 50\% of data continuously in a set 60 minutes of time series for ARIMA to impute. We choose 15 minutes since we observe that ARIMA yields significantly higher mean absolute errors when the missing data rate is 50\% (30 minutes) than for missing rates at 25\% (15 minutes) and 10\% (6 minutes).

\noindent\textbf{Data Filtering} Wearable sensor recordings are vulnerable to noise and motion artifacts and thus filtering becomes an essential prerequisite for further processing of the signal. We address this issue by applying the Savitzky-Golay filter \cite{s_g_savitzky}. This is a well-known smoothing approach that increases the precision of signals while simultaneously preserving the signal trend. The filter tries to fit a polynomial with some pre-selected fixed degree $z$ to the time-series sequence $\mathbf{S}$ of length $2m + 1$ centered at $i = 0$ such that the squared error is minimized over coefficients of a polynomial $p_{i} = \sum_{x=0}^{z} a_{x}i^{x}$:

\vspace{-2mm}
\begin{equation*}
    \min_{a_{0:z}} \sum_{i = -m} ^ {m} (p_{i} - s_{i}) ^ 2
\end{equation*}

It is worth noting that there are many other attractive approaches available in the literature to pre-process the time-series data, but empirically testing each is beyond our scope in this work.

\subsection{Time-series Clustering}
\noindent\textbf{Definitions} We first introduce some notations and definitions used in this and later sections. The pre-processed time-series data is defined as a set of observations, $(\mathbf{s}_1, \mathbf{s}_2,..., \mathbf{s}_{n})$ ordered by time, where each $\mathbf{s}_i \in \mathit{R}^m$ is the $i$-th observation in time with $m$ features. The sensor features in this case are PPG heart rate and step-count that are observed every minute, so $m=2$. We further create $\mathbf{X}_{i}$ which consists of $w$ observations $(\mathbf{s}_i, \mathbf{s}_{i+1}, ... , \mathbf{s}_{i+w-1})$. The aim of TICC described next is to partition these sequential time-series observations to form $K$ clusters.

\noindent\textbf{Toeplitz Inverse Covariance-Based Clustering (TICC)} In this method, each cluster is characterized by a Toeplitz Gaussian inverse covariance $\Theta_{k} \in \mathit{R}^{mw\times mw}$ and empirical mean $\mu^{m}$. The Toeplitz Gaussian inverse covariance essentially captures the interdependencies between different observations. This clustering approach also enforces temporal consistency between consecutive vectors $\mathbf{X}_{i}$ and $\mathbf{X}_{i+1}$ to find repeated long-range patterns in the data that represent particular behaviors of the object. In summary, the TICC method assigns each frame to one Gaussian inverse covariance $\Theta_{k}$ by minimizing the following objective function:

\vspace{-1mm}
\begin{equation*}
    \underset{\mathbf{P}, \Theta}{\text{minimize}} \sum_{k=1}^{K} \sum_{\mathbf{X_{i}} \in \mathbf{P}_{k}} (-\mathit{ll}(\mathbf{X}_{i}, \Theta_{k}) + \beta \cdot \mathds{1}_{\mathbf{X}_{i} \not\in \mathbf{P}_{k}})
    \label{equ:ticc}
\end{equation*}

where in Eq.~\ref{equ:ticc}, $K$ is the number of clusters and $\mathbf{P}_{k}$ corresponds to the cluster point set that belongs to cluster $k$. $\mathds{1}_{\mathbf{X}_{j} \not\in \mathbf{P}_{k}}$ is an indicator function equal to one when the current cluster $\mathbf{X}_{j}$ is different from the future cluster assignment and zero otherwise. $\beta$ is the penalty parameter that controls the temporal consistency. A larger $\beta$ will result in encouraging the neighboring samples to be assigned to the same cluster. \cite{hallac2017toeplitz} also suggests that the performance of time-series clustering largely depends on the choice of $\beta$ when the sample size is adequate. Finally, $\mathit{ll}(\mathbf{X}_{j}, \Theta_{k})$ represents the log-likelihood that $\mathbf{X_{j}}$ belongs to $\mathbf{P}_{k}$, which is defined below:

\begin{equation}
    \begin{aligned}
         -\mathit{ll}(\mathbf{X}_{i}, \Theta_{k}) = -\frac{1}{2}(\mathbf{X}_{i} - \mathbf{\mu_{k}})^\intercal \Theta_{k}(\mathbf{X}_{i} - \mathbf{\mu_{k}}) 
         \\ + \frac{1}{2}\log{\det\Theta_{k}} - \frac{m}{2}\log(2\pi)
    \end{aligned}
    \label{equ:ll}
\end{equation}

where $\mathbf{\mu_{k}}$ is the empirical mean of cluster $k$, and $m$ is the number of features in each observations. In our context, we choose this method for time-series clustering since our interest is to robustly identify long-range repeated patterns while contending with the possible presence of irrelevant data points.

\subsection{Hawkes Point Process}

Hawkes processes \cite{hawkes_hawkes}, also known as a self-exciting counting process, is a stochastic point process for studying event sequence patterns where historical event occurrences are assumed to increase the chance of arrival of new events. In general, a point process is a collection of a list of discrete events in time and their associate dimension, $\{t_{i}, u_{i}\}$ with time $t_{i} \in [0, n]$ and event $u_{i} \in [0, U]$, where $n$ and $U$ are the maximum time in a time-series and total types of events, respectively. Here, we define $d_{i}$ representing the clusters that assign to $i$-th time point. In this study, we define the event as transitions between different time-series clusters. For instance, given two possible consecutive time points $\{t_{i}, d_{i}\}$ and $\{t_{i+1}, d_{i+1}\}$ in a time-series of a day, we can define an event as $\{t_{i+1}, (d_{i} \rightarrow d_{i+1})\}$, such that $d_{i} \neq d_{i+1}$. In this work, we can also find that $n$ represents the number of total points in a time-series of days.

A multi-dimensional point process with $U$ types of events can be equivalently represented by $\mathcal{U}$ counting processes: $\mathcal{N}_{u} = \{\mathcal{N}_{u}(t)| t \in [0, n]\}$, where $\mathcal{N}_{u}(t)$ denotes the number of type-$u$ events occurring before $n$-th time point. There are possible $\binom{K}{2}-K$ types of events by the definition of the event in this work. Let the history $\mathcal{H}(t)$ be the list of times of events up to time $n$:

\begin{equation}
    \mathcal{H}^{\mathcal{U}}_{n} = \{ (t_{i}, u_{i}) | t_{i} < n, u_{i} \in \mathcal{U} \}
    \label{equ:history}
\end{equation}

\noindent Then, the expected instantaneous rate of occurring type-$u$ events given history is :
\begin{equation}
    \mathcal{\lambda}_{u}(t) dt = \mathcal{\lambda}_{u}(t|\mathcal{H}^\mathcal{U}_{t})
\end{equation}

\noindent We can then derive the intensity function $\mathcal{\lambda}_{u}(t)$ of the multi-dimensional Hawkes process as:
\begin{equation}
    \begin{aligned}
        \mathcal{\lambda}_{u}(t) &= \mu_{u} + \sum_{i:t_{i}<t} \phi_{uu'}(t_{i})
    \end{aligned}
\end{equation}

\noindent where $\mu_{u}$ is the exogenous (base) intensity indecent of the history. $\phi_{uu'}(t)$ is the impact function capturing the temporal influence of type-$u'$ event on the subsequent type-$u$ event. We can further define $\phi_{uu'}(t)$ as:

\vspace{-1mm}
\begin{equation}
    \begin{aligned}
        \phi_{uu'}(t) = \sum_{m=1}^{M} a_{uu'}^{m} k_{m}(t)
    \end{aligned}
\end{equation}

\noindent where $k_{m}(t)$ is the m-th basis function and $a_{uu'}^{m}$ represents the coefficient corresponding to $k_{m}(t)$. We adopted the Gaussian filter as the basis function in this study. We describe type-$u'$ event does not Granger-cause type-$u$ event, when function $\lambda _{u}(t)$ is independent of historical events of type-$u'$. Finally, we can build a Granger causality graph $G=(\mathcal{U}, \mathcal{E})$ with $U$ types of events as the nodes and the directed edges in between representing the causation. In this study, the routine feature is naturally defined as the adjacency matrix $\mathbf{A}$ of the Granger causality graph learned from the Hawkes process. Element $\mathbf{A}_{uu'}$ can be viewed as the infectivity ($\int_{0}^{t}\phi_{uu'}(s) ds$) of type-$u'$ cluster-transition on type-$u$ cluster-transition. 

A detailed description of the above definitions can be found in \cite{gc_hawkes, hawkes_xu}. To discover the Granger causality graph from a Hawkes point process, we adopt the learning algorithm proposed by \cite{hawkes_xu}. This algorithm learns the Granger causality graph robustly given a few training sequences. 

\noindent\textbf{Summary} Our proposed HOT-ROD pipeline aims to learn routine characteristics from wearable time-series without prior knowledge or the acquisition of labeled event sequences. The time series data used in this study contains continuous measurements of physiological response and physical activity. The proposed learned routine features capture patterns in how people advance from one state to another in everyday life.

\section{Result}
\label{sec:experiment}

In this section, we validate the effectiveness of the proposed HOT-ROD approach for predicting demographic, personality and affect with Fitbit time series data.

\begin{table*}[ht]

    \centering
    \small
    \caption{Table shows quantitative comparison (macro-F1 score) of different routine features to predict \texttt{Initial Assessment} survey using Random Forest Model}
    \def\arraystretch{1.28}
    \begin{tabular}{ccccccccc}
        \hline
        & \multicolumn{1}{c}{Clusters} & \multicolumn{1}{c}{Neuroticism} & \multicolumn{1}{c}{Conscientiousness} & \multicolumn{1}{c}{Extraversion} & \multicolumn{1}{c}{Agreeableness} & \multicolumn{1}{c}{Openness} & \multicolumn{1}{c}{Pos-Affect} & \multicolumn{1}{c}{Neg-Affect} \\ \hline
        Fitbit Summary & \multicolumn{1}{c}{-} &
        \multicolumn{1}{c}{60.85\%} & \multicolumn{1}{c}{49.84\%} & \multicolumn{1}{c}{64.91\%} & \multicolumn{1}{c}{63.95\%} & \multicolumn{1}{c}{60.86\%} & 
        \multicolumn{1}{c}{58.24\%} & \multicolumn{1}{c}{64.88\%} \\ \hline
        
        \multirow{3}{*}{HOT-ROD}  & 3 &
        \multicolumn{1}{c}{63.23\%} & \multicolumn{1}{c}{66.60\%} & \multicolumn{1}{c}{62.12\%} & \multicolumn{1}{c}{63.64\%} & \multicolumn{1}{c}{\textbf{62.19\%}} & 
        \multicolumn{1}{c}{54.32\%} & \multicolumn{1}{c}{62.82\%} \\
         & 4 &
        \multicolumn{1}{c}{{56.21\%}} & \multicolumn{1}{c}{\textbf{68.08\%}} & \multicolumn{1}{c}{62.64\%} & \multicolumn{1}{c}{63.85\%} & \multicolumn{1}{c}{54.08\%} & 
        \multicolumn{1}{c}{54.39\%} & \multicolumn{1}{c}{58.00\%}  \\
         & 5 &
        \multicolumn{1}{c}{61.07\%} & \multicolumn{1}{c}{{60.29\%}} & \multicolumn{1}{c}{58.53\%} & \multicolumn{1}{c}{64.00\%} & \multicolumn{1}{c}{{57.83\%}} & 
        \multicolumn{1}{c}{62.48\%} & \multicolumn{1}{c}{58.30\%}  \\ \hline
        
        \multirow{3}{*}{HOT-ROD + Fitbit Summary}  & 3 &
        \multicolumn{1}{c}{\textbf{63.53\%}} & \multicolumn{1}{c}{52.09\%} & \multicolumn{1}{c}{\textbf{66.25\%}} & \multicolumn{1}{c}{\textbf{70.13\%}} & \multicolumn{1}{c}{{60.86\%}} & 
        \multicolumn{1}{c}{58.35\%} & \multicolumn{1}{c}{\textbf{65.72\%}} \\
         & 4 &
        \multicolumn{1}{c}{59.84\%} & \multicolumn{1}{c}{{63.65\%}} & \multicolumn{1}{c}{{65.87\%}} & \multicolumn{1}{c}{{64.76\%}} & \multicolumn{1}{c}{{56.09\%}} & 
        \multicolumn{1}{c}{{59.35\%}} & \multicolumn{1}{c}{61.91\%}  \\
         & 5 &
        \multicolumn{1}{c}{60.56\%} & \multicolumn{1}{c}{58.03\%} & \multicolumn{1}{c}{65.02\%} & \multicolumn{1}{c}{64.40\%} & \multicolumn{1}{c}{59.58\%} & 
        \multicolumn{1}{c}{\textbf{63.67\%}} & \multicolumn{1}{c}{{60.31\%}}  \\ \hline
    \end{tabular}
    \label{tab:result}
\end{table*}

\begin{table}[ht]

    \centering
    \caption{Table shows quantitative comparison (macro-F1 score) of different routine features to predict \texttt{demographics} using Random Forest Model.}
    \def\arraystretch{1.25}
    \begin{tabular}{cccc}
        \hline
        & \multicolumn{1}{c}{Clusters} & \multicolumn{1}{c}{Work Shift} & \multicolumn{1}{c}{Job Type} \\ \hline
        Fitbit Summary & \multicolumn{1}{c}{-} & \multicolumn{1}{c}{68.92\%} & \multicolumn{1}{c}{60.55\%} \\ \hline
        
        \multirow{3}{*}{HOT-ROD} & 3 & \multicolumn{1}{c}{58.76\%} & \multicolumn{1}{c}{68.79\%} \\ 
         & 4 & \multicolumn{1}{c}{55.98\%} & \multicolumn{1}{c}{67.42\%}  \\
         & 5 & \multicolumn{1}{c}{62.68\%} & \multicolumn{1}{c}{70.13\%}  \\ \hline
         
        \multirow{3}{*}{HOT-ROD + Fitbit Summary} & 3 & \multicolumn{1}{c}{\textbf{69.16\%}} & \multicolumn{1}{c}{62.50\%} \\ 
         & 4 & \multicolumn{1}{c}{{64.64\%}} & \multicolumn{1}{c}{{64.61\%}}  \\
         & 5 & \multicolumn{1}{c}{65.99\%} & \multicolumn{1}{c}{\textbf{70.13\%}}  \\ \hline
        
    \end{tabular}
    \label{tab:demographic}
    \vspace{-3mm}
\end{table}

\subsection{Experimental Setup}

\subsubsection{Daily Summary Routine Feature}

We use Fitbit daily summary data to construct our baseline model. The Fitbit daily summary data is extracted using the API provided by Fitbit. The measurements we select from Fitbit daily summary report include sleep duration, sleep efficiency, step counts, resting heart rate, and heart rate zone duration. Here, Fitbit categorizes heart rate into $4$ zones: 
\begin{itemize}[leftmargin=*]
    \item Out of Zone (heart rate is below $50\%$ of its maximum).
    \item Fat-burn Zone (heart rate is $51\%$ to $69\%$ to its maximum).
    \item Cardio Zone (heart rate is $70\%$ to $84\%$ to its maximum).
    \item Peak Zone (heart rate is above $85\%$ of its maximum).
\end{itemize}

A set of 5 statistical functionals (e.g., max, standard deviation) are then introduced on the Fitbit summary feature. 


\subsubsection{HOT-ROD Routine Feature}

We compute the HOT-ROD routine features from Fitbit Charge $2$ time-series that measure PPG heart rate and step count. Routine behaviors may differ between workdays and off-days. Thus we separately learn the routine features of workdays and off-days. We also observe that the data available for each participant varies (min: $2$ days, max: $70$ days) which can lead to biased results without proper data selection. Hence, we choose to randomly select an equal amount of days of data from workdays and off-days of a participant for our analysis. In our study, we experimentally test by picking $n=5$ days from each type. We choose to pick $5$ days in our experiment since it ensures a reasonable amount of data per participant and also the number of qualified participants retained in the analysis. In the end, there are 101 participants retained in this experiment.

According to Section~\ref{sec:method}, we first aggregate PPG heart rate and step count every minute for a day of data. We impute the missing data in each aggregated time-series using the ARIMA model, where we estimate the number of time lags, the degree of difference, and the order of the moving average according to the Akaike information criterion. We fill the continuous missing segments over 15 minutes with large enough negative values. We then filter the imputed time series using an S-G filter. To minimize the deforming of the signal, we choose a small window size $m = 2$, and cubic order polynomials in the S-G filter. The window size parameter could be tuned systematically based on criteria and heuristics defined in \cite{vivo2006automatic}, but we leave this endeavor for future work.

Prior to time-series clustering, we perform the z-normalization of each time-series to remove variances between participants. We empirically experiment with the number of clusters $K =\{3, 4, 5\}$ in TICC. We set $\beta$ to be 10 in this experiment to encourage neighboring samples to be assigned to the same cluster. We want to highlight that we plan to systematically tune $\beta$ in our future works. Since some input time series may still contain portions of the missing elements (the segment that is missing above 15 minutes), one cluster output is associated with "missing measurements". We decide to ignore this cluster in the following analysis procedure. Finally, we applied the Hawkes Point Process to learn the infectivity between cluster transitions.

\begin{figure*}[ht]
	\centering
	\includegraphics[width=\textwidth]{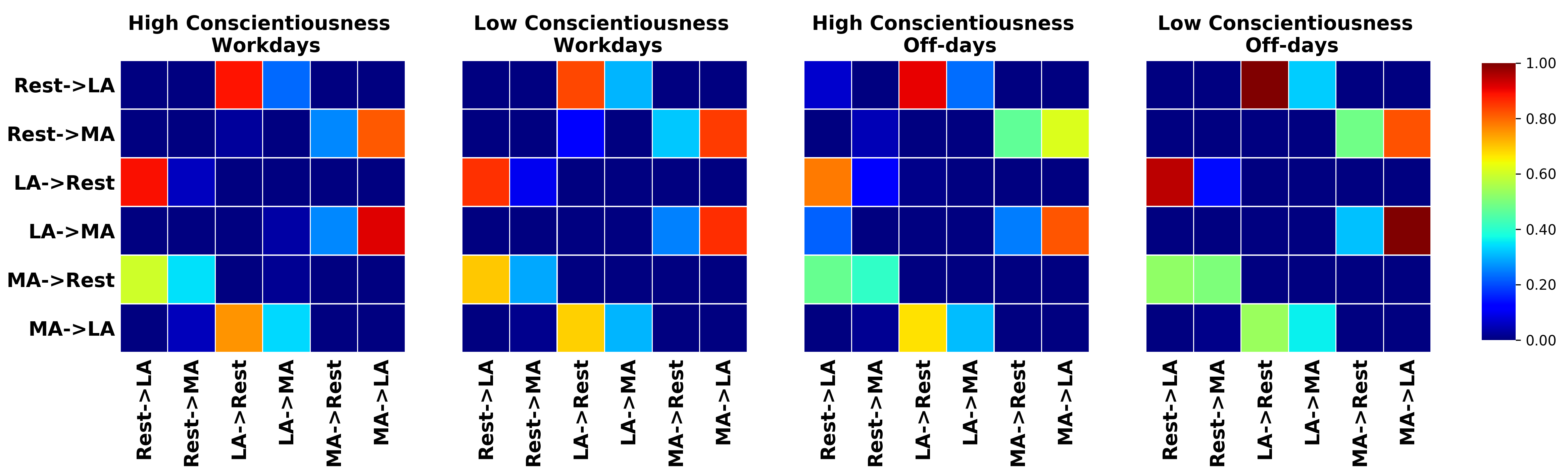}
    \caption{Figure shows the infectivity matrix for various cluster-transition types in the high level of conscientiousness and low level of conscientiousness population, with a higher value representing stronger causal relations between $2$ cluster-transition events. LA and MA represent light activity and moderate activity, respectively.}
    \label{fig:con_infectivity}
\end{figure*}

\subsubsection{Model Description}

We observe that the distributions of ground truth assessments exhibited significant skew, posing a difficulty for most supervised learning algorithms, as they will be biased towards the majority group, leading to poor predictions on minority labels. Therefore, we choose to binarize the ground truth label as our final prediction target. We binarize personality, affect scores by the median split, while we categorize job type and work shift as nurses/non-nurses and day-shift/night shift, respectively. We then evaluate the efficacy of the Fitbit summary routine feature and HOT-ROD routine features extracted above by predicting binarized ground-truth labels using the Random Forest (RF) classifier. We select Random Forest models since they have considerable advantages over other techniques with respect to robustness to noise, tuning simplicity, and the ability to choose the most relevant features from high-dimensional data input, where many features are often redundant \cite{random_forest_feat_selection}. Specifically, we perform the predictions using three sets of features: 1. Fitbit summary routine feature; 2. HOT-ROD routine feature; 3. Fitbit summary routine feature and HOT-ROD routine feature. We perform 5-fold cross-validation and report the average results in the macro-F1 score. We grid search the hyper-parameters in the RF model as follows: 1. Number of estimators: [10, 20, 30]; 2. Feature selection criterion: ["gini", "entropy"]; 3. Max depth of the tree: [4, 5, 6]; 4. Minimum samples to split: [2, 3, 5].

\subsection{Prediction Results}
The experimental results for predicting IGTB assessment (personality and affect) and demographic information (job type, shift) are listed in Table \ref{tab:result} and Table \ref{tab:demographic} respectively. For predicting demographics, HOT-ROD features combined with Fitbit summary routine features achieve the best performance, with a better F1 score in predicting work shift than only using either the Fitbit summary feature or HOT-ROD routine features. We also observe that HOT-ROD reaches the best performance in forecasting neuroticism when the assigned number of clusters is $4$. HOT-ROD features can also outperform Fitbit summary and combined feature set in classifying conscientiousness and openness when number of cluster is 3. Combining feature sets achieves the best performance in determining extraversion, agreeableness, and affect related variables.

\section{Discussions}
\label{sec:discussion}

The results in Table \ref{tab:result} and Table \ref{tab:demographic} demonstrate foremost that personality and affect prediction from this data (collected in a natural setting outside of a well-controlled lab) is considerably challenging using standard physiologic features and simple machine learning techniques since none of the validation scores can reach above 70\% even when the label is binarized. It also suggests that routine patterns, derived from wearable recordings by our HOT-ROD analysis, modestly improve performance. The prediction results demonstrate the HOT-ROD feature work comparatively reliably when the number of clusters is below 4. This may be because the available points for each cluster transition in a day decrease dramatically when the number of clusters increases, leading to an imprecise estimation of the Granger graph from time-series event.

We further identify that HOT-ROD features yield substantially better performances in predicting conscientiousness, which is known for measuring self-discipline. Increasingly, HOT-ROD routine features combined with Fitbit summary routines are better predictors of job type and shift type. These prediction results demonstrate that HOT-ROD routine features are aligned quite well with the job type and self-discipline which closely relate to human behavior. In the end, we observe that the HOT-ROD features to produce a noticeably better F1-score in classifying positive affect. Though it is easy to find that the HOT-ROD pipeline can capture routine features to predict human behaviors using only a few days of data, the prediction results are related to the number of clusters assigned in TICC. It draws attention to the fact that more data are needed to generate a reliable estimation of routine behavior.

To better understand the learned routine features in the case when the number of clusters $K = 4$, we further infer an interpretation for each cluster as follows: 1. Rest activity; 2. Light activity; 3. Moderate activity or exercises; 4. Missing measurements.

Fig~\ref{fig:con_infectivity} displays the infectivity matrix for various cluster transitions of high and low conscientious groups. There are 13 noticeable causality relations between cluster transitions, while none of the cluster-transition events have obvious self-triggering patterns. This implies that most events don't have a periodic daily behavior. Both groups behaved similarly on workdays, while \textbf{Light activity} $\rightarrow$ \textbf{Rest} transitions are more likely to be triggered by \textbf{Rest} $\rightarrow$ \textbf{Light activity} transitions in low conscientious population on off-days. Additionally, \textbf{Rest} $\rightarrow$ \textbf{Moderate activity} transitions also tend to impact \textbf{Moderate activity} $\rightarrow$ \textbf{Light activity} transitions in low conscientious population on off-days. These causal relations indicate that the low conscientious population tends to be more sedentary and less active on days of not working. This finding is consistent with the prior work stated in \cite{con_sed_allen} and \cite{con_sed_sutin}. Consistent with the feature importance returned from the random forest model, we also observe that these learned causal relations offer important information in predicting conscientious type. Similarly, we also observe that the \textbf{Rest} $\rightarrow$ \textbf{Moderate activity} is more likely to be triggered by the \textbf{Moderate activity} $\rightarrow$ \textbf{Rest} on off-days in the group with the high level of positive affect scores.

\section{Conclusion}
\label{sec:conclusion}
We propose a technique, HOT-ROD, for discovering routine patterns in wearable sensor time-series data utilizing the Granger causality graph extracted from a time-series of cluster-transition events. Using a data set of over $100$ participants working in a hospital environment for ten weeks, we show that this data-driven technique intuitively captures transitional behaviors between activity states in a manner consistent with personality without using any prior knowledge. We have also shown that routine features extracted from HOT-ROD combined with routine features derived using Fitbit daily summary information modestly improve the performance in predicting job type, work shift, extraversion, agreeableness and affect variables than using a single set of features. 

As the next step, we want to examine how the switching penalty $\beta$ impacts the learned routine behaviors. Increasingly, we believe the performance of the proposed technique will further increase when more data is available, but we also hope to systematically study the amount of data required to learn the robust routine behavior patterns using our approach. In addition, we believe this technique is simple enough to generalize to other data sets, possibly more broadly than wearable sensor readings. Ultimately, we plan to compare our method with other popular time-series approaches presented using time-series clustering \cite{paparrizos2015k, paparrizos2017fast}, motif finding \cite{yeh2016matrix, feng2019discovering}, and deep representation learning \cite{ekambaram2023tsmixer, yue2022ts2vec, yu2023semi, zhang2023self}.

\section{Acknowledgement}

This research is based upon work supported by the Office of the Director of National Intelligence (ODNI), Intelligence Advanced Research Projects Activity (IARPA), via IARPA Contract No. 2017-17042800005. The views and conclusions contained herein are those of the authors and should not be interpreted as necessarily representing the official policies or endorsements, either expressed or implied, of the ODNI, IARPA, or the U.S. Government. The U.S. Government is authorized to reproduce and distribute reprints for Governmental purposes notwithstanding any copyright annotation thereon.

\bibliographystyle{ACM-Reference-Format}
\bibliography{ref}


\begin{thebibliography}{50}


\ifx \showCODEN    \undefined \def \showCODEN     #1{\unskip}     \fi
\ifx \showDOI      \undefined \def \showDOI       #1{#1}\fi
\ifx \showISBNx    \undefined \def \showISBNx     #1{\unskip}     \fi
\ifx \showISBNxiii \undefined \def \showISBNxiii  #1{\unskip}     \fi
\ifx \showISSN     \undefined \def \showISSN      #1{\unskip}     \fi
\ifx \showLCCN     \undefined \def \showLCCN      #1{\unskip}     \fi
\ifx \shownote     \undefined \def \shownote      #1{#1}          \fi
\ifx \showarticletitle \undefined \def \showarticletitle #1{#1}   \fi
\ifx \showURL      \undefined \def \showURL       {\relax}        \fi
\providecommand\bibfield[2]{#2}
\providecommand\bibinfo[2]{#2}
\providecommand\natexlab[1]{#1}
\providecommand\showeprint[2][]{arXiv:#2}

\bibitem[{A. Roebuck and V. Monasterio and E. Gederi and M. Osipov and J.
  Behar, A. Malhotra and T. Penzel and G. D. Clifford}(2014)]%
        {sleep_roebuck}
\bibfield{author}{\bibinfo{person}{{A. Roebuck and V. Monasterio and E. Gederi
  and M. Osipov and J. Behar, A. Malhotra and T. Penzel and G. D. Clifford}}.}
  \bibinfo{year}{2014}\natexlab{}.
\newblock \showarticletitle{A review of signals used in sleep analysis}.
\newblock \bibinfo{journal}{\emph{Physiological Measurement}}
  \bibinfo{volume}{35}, \bibinfo{number}{1} (\bibinfo{year}{2014}),
  \bibinfo{pages}{R1}.
\newblock


\bibitem[{A. Savitzky and M. Golay}(1964)]%
        {s_g_savitzky}
\bibfield{author}{\bibinfo{person}{{A. Savitzky and M. Golay}}.}
  \bibinfo{year}{1964}\natexlab{}.
\newblock \showarticletitle{Smoothing and Differentiation of Data by Simplified
  Least Squares Procedures}.
\newblock \bibinfo{journal}{\emph{Analytical chemistry}}  \bibinfo{volume}{36}
  (\bibinfo{date}{July} \bibinfo{year}{1964}), \bibinfo{pages}{1627--1639}.
\newblock


\bibitem[Allen(2007)]%
        {ppg_clinical}
\bibfield{author}{\bibinfo{person}{J. Allen}.} \bibinfo{year}{2007}\natexlab{}.
\newblock \showarticletitle{Photoplethysmography and its application in
  clinical physiological measurement}.
\newblock \bibinfo{journal}{\emph{Physiological Measurement}}
  \bibinfo{volume}{28} (\bibinfo{date}{Jan.} \bibinfo{year}{2007}).
\newblock


\bibitem[Allen et~al\mbox{.}(2016)]%
        {con_sed_allen}
\bibfield{author}{\bibinfo{person}{M. Allen}, \bibinfo{person}{E. Walter},
  {and} \bibinfo{person}{M. McDermott}.} \bibinfo{year}{2016}\natexlab{}.
\newblock \showarticletitle{Personality and Sedentary Behavior: A Systematic
  Review and Meta-Analysis}.
\newblock \bibinfo{journal}{\emph{Health Psychology}}  \bibinfo{volume}{36}
  (\bibinfo{date}{10} \bibinfo{year}{2016}).
\newblock


\bibitem[Avramidis et~al\mbox{.}(2022)]%
        {avramidis2022multimodal}
\bibfield{author}{\bibinfo{person}{Kleanthis Avramidis},
  \bibinfo{person}{Tiantian Feng}, \bibinfo{person}{Digbalay Bose}, {and}
  \bibinfo{person}{Shrikanth Narayanan}.} \bibinfo{year}{2022}\natexlab{}.
\newblock \showarticletitle{Multimodal Estimation of Change Points of
  Physiological Arousal in Drivers}.
\newblock \bibinfo{journal}{\emph{arXiv preprint arXiv:2210.15826}}
  (\bibinfo{year}{2022}).
\newblock


\bibitem[Baldwin(1988)]%
        {routine_baldwin}
\bibfield{author}{\bibinfo{person}{J.D. Baldwin}.}
  \bibinfo{year}{1988}\natexlab{}.
\newblock \showarticletitle{Habit, Emotion, and Self-Conscious Action}.
\newblock \bibinfo{journal}{\emph{Sociological Perspectives}}
  \bibinfo{volume}{31}, \bibinfo{number}{1} (\bibinfo{year}{1988}),
  \bibinfo{pages}{35--57}.
\newblock


\bibitem[Booth et~al\mbox{.}(2019a)]%
        {booth2019toward}
\bibfield{author}{\bibinfo{person}{Brandon~M Booth}, \bibinfo{person}{Tiantian
  Feng}, \bibinfo{person}{Abhishek Jangalwa}, {and}
  \bibinfo{person}{Shrikanth~S Narayanan}.} \bibinfo{year}{2019}\natexlab{a}.
\newblock \showarticletitle{Toward robust interpretable human movement pattern
  analysis in a workplace setting}. In \bibinfo{booktitle}{\emph{ICASSP
  2019-2019 IEEE International Conference on Acoustics, Speech and Signal
  Processing (ICASSP)}}. IEEE, \bibinfo{pages}{7630--7634}.
\newblock


\bibitem[Booth et~al\mbox{.}(2019b)]%
        {booth2019multimodal}
\bibfield{author}{\bibinfo{person}{Brandon~M Booth}, \bibinfo{person}{Karel
  Mundnich}, \bibinfo{person}{Tiantian Feng}, \bibinfo{person}{Amrutha
  Nadarajan}, \bibinfo{person}{Tiago~H Falk}, \bibinfo{person}{Jennifer~L
  Villatte}, \bibinfo{person}{Emilio Ferrara}, {and} \bibinfo{person}{Shrikanth
  Narayanan}.} \bibinfo{year}{2019}\natexlab{b}.
\newblock \showarticletitle{Multimodal human and environmental sensing for
  longitudinal behavioral studies in naturalistic settings: Framework for
  sensor selection, deployment, and management}.
\newblock \bibinfo{journal}{\emph{Journal of medical Internet research}}
  \bibinfo{volume}{21}, \bibinfo{number}{8} (\bibinfo{year}{2019}),
  \bibinfo{pages}{e12832}.
\newblock


\bibitem[Box and Jenkins(1990)]%
        {arima}
\bibfield{author}{\bibinfo{person}{G. Box} {and} \bibinfo{person}{G. Jenkins}.}
  \bibinfo{year}{1990}\natexlab{}.
\newblock \bibinfo{booktitle}{\emph{Time Series Analysis, Forecasting and
  Control}}.
\newblock \bibinfo{publisher}{Holden-Day, Inc.}
\newblock


\bibitem[Che et~al\mbox{.}(2018)]%
        {che2018recurrent}
\bibfield{author}{\bibinfo{person}{Zhengping Che}, \bibinfo{person}{Sanjay
  Purushotham}, \bibinfo{person}{Kyunghyun Cho}, \bibinfo{person}{David
  Sontag}, {and} \bibinfo{person}{Yan Liu}.} \bibinfo{year}{2018}\natexlab{}.
\newblock \showarticletitle{Recurrent neural networks for multivariate time
  series with missing values}.
\newblock \bibinfo{journal}{\emph{Scientific reports}} \bibinfo{volume}{8},
  \bibinfo{number}{1} (\bibinfo{year}{2018}), \bibinfo{pages}{6085}.
\newblock


\bibitem[Dawadi et~al\mbox{.}(2016)]%
        {life_seq_dawadi}
\bibfield{author}{\bibinfo{person}{P.N. Dawadi}, \bibinfo{person}{D.J. Cook},
  {and} \bibinfo{person}{M. Schmitter-Edgecombe}.}
  \bibinfo{year}{2016}\natexlab{}.
\newblock \showarticletitle{Modeling patterns of activities using activity
  curves}.
\newblock \bibinfo{journal}{\emph{Pervasive and Mobile Computing}}
  \bibinfo{volume}{28} (\bibinfo{year}{2016}), \bibinfo{pages}{51 -- 68}.
\newblock


\bibitem[Didelez(2008)]%
        {gc_hawkes}
\bibfield{author}{\bibinfo{person}{V. Didelez}.}
  \bibinfo{year}{2008}\natexlab{}.
\newblock \showarticletitle{Graphical models for marked point processes based
  on local independence}.
\newblock \bibinfo{journal}{\emph{Journal of the Royal Statistical Society
  Series B}}  \bibinfo{volume}{70} (\bibinfo{date}{February}
  \bibinfo{year}{2008}), \bibinfo{pages}{245--264}.
\newblock


\bibitem[Dong and Peng(2013)]%
        {imputation_dong}
\bibfield{author}{\bibinfo{person}{Y. Dong} {and} \bibinfo{person}{J. Peng}.}
  \bibinfo{year}{2013}\natexlab{}.
\newblock \showarticletitle{Principled missing data methods for researchers}.
\newblock \bibinfo{journal}{\emph{SpringerPlus}}  \bibinfo{volume}{2}
  (\bibinfo{date}{December} \bibinfo{year}{2013}), \bibinfo{pages}{222}.
\newblock


\bibitem[Ekambaram et~al\mbox{.}(2023)]%
        {ekambaram2023tsmixer}
\bibfield{author}{\bibinfo{person}{Vijay Ekambaram}, \bibinfo{person}{Arindam
  Jati}, \bibinfo{person}{Nam Nguyen}, \bibinfo{person}{Phanwadee Sinthong},
  {and} \bibinfo{person}{Jayant Kalagnanam}.} \bibinfo{year}{2023}\natexlab{}.
\newblock \showarticletitle{TSMixer: Lightweight MLP-Mixer Model for
  Multivariate Time Series Forecasting}.
\newblock \bibinfo{journal}{\emph{arXiv preprint arXiv:2306.09364}}
  (\bibinfo{year}{2023}).
\newblock


\bibitem[Endo et~al\mbox{.}(2016)]%
        {trajectory_endo}
\bibfield{author}{\bibinfo{person}{Y. Endo}, \bibinfo{person}{H. Toda},
  \bibinfo{person}{K. Nishida}, {and} \bibinfo{person}{J. Ikedo}.}
  \bibinfo{year}{2016}\natexlab{}.
\newblock \showarticletitle{Classifying spatial trajectories using
  representation learning}.
\newblock \bibinfo{journal}{\emph{International Journal of Data Science and
  Analytics}} \bibinfo{volume}{2}, \bibinfo{number}{3} (\bibinfo{date}{01 Dec}
  \bibinfo{year}{2016}), \bibinfo{pages}{107--117}.
\newblock


\bibitem[Fahad et~al\mbox{.}(2013)]%
        {home_setting_neural}
\bibfield{author}{\bibinfo{person}{L.G. Fahad}, \bibinfo{person}{A. Ali},
  \bibinfo{person}{}, {and} \bibinfo{person}{M. Rajarajan}.}
  \bibinfo{year}{2013}\natexlab{}.
\newblock \showarticletitle{Long term analysis of daily activities in smart
  home}. In \bibinfo{booktitle}{\emph{ESANN}}.
\newblock


\bibitem[Feng et~al\mbox{.}(2021)]%
        {feng2021multimodal}
\bibfield{author}{\bibinfo{person}{Tiantian Feng}, \bibinfo{person}{Brandon~M
  Booth}, \bibinfo{person}{Brooke Baldwin-Rodr{\'\i}guez},
  \bibinfo{person}{Felipe Osorno}, {and} \bibinfo{person}{Shrikanth
  Narayanan}.} \bibinfo{year}{2021}\natexlab{}.
\newblock \showarticletitle{A multimodal analysis of physical activity, sleep,
  and work shift in nurses with wearable sensor data}.
\newblock \bibinfo{journal}{\emph{Scientific reports}} \bibinfo{volume}{11},
  \bibinfo{number}{1} (\bibinfo{year}{2021}), \bibinfo{pages}{8693}.
\newblock


\bibitem[Feng et~al\mbox{.}(2018)]%
        {feng2018tiles}
\bibfield{author}{\bibinfo{person}{Tiantian Feng}, \bibinfo{person}{Amrutha
  Nadarajan}, \bibinfo{person}{Colin Vaz}, \bibinfo{person}{Brandon Booth},
  {and} \bibinfo{person}{Shrikanth Narayanan}.}
  \bibinfo{year}{2018}\natexlab{}.
\newblock \showarticletitle{Tiles audio recorder: an unobtrusive wearable
  solution to track audio activity}. In \bibinfo{booktitle}{\emph{Proceedings
  of the 4th ACM Workshop on Wearable Systems and Applications}}.
  \bibinfo{pages}{33--38}.
\newblock


\bibitem[Feng and Narayanan(2019a)]%
        {feng2019imputing}
\bibfield{author}{\bibinfo{person}{Tiantian Feng} {and}
  \bibinfo{person}{Shrikanth Narayanan}.} \bibinfo{year}{2019}\natexlab{a}.
\newblock \showarticletitle{Imputing missing data in large-scale multivariate
  biomedical wearable recordings using bidirectional recurrent neural networks
  with temporal activation regularization}. In \bibinfo{booktitle}{\emph{2019
  41st Annual International Conference of the IEEE Engineering in Medicine and
  Biology Society (EMBC)}}. IEEE, \bibinfo{pages}{2529--2534}.
\newblock


\bibitem[Feng and Narayanan(2019b)]%
        {feng2019discovering}
\bibfield{author}{\bibinfo{person}{Tiantian Feng} {and}
  \bibinfo{person}{Shrikanth~S Narayanan}.} \bibinfo{year}{2019}\natexlab{b}.
\newblock \showarticletitle{Discovering optimal variable-length time series
  motifs in large-scale wearable recordings of human bio-behavioral signals}.
  In \bibinfo{booktitle}{\emph{ICASSP 2019-2019 IEEE International Conference
  on Acoustics, Speech and Signal Processing (ICASSP)}}. IEEE,
  \bibinfo{pages}{7615--7619}.
\newblock


\bibitem[Halberg et~al\mbox{.}(1977)]%
        {circadian_halberg}
\bibfield{author}{\bibinfo{person}{F. Halberg}, \bibinfo{person}{F.
  Carandente}, \bibinfo{person}{G. Cornelissen}, {and} \bibinfo{person}{G.S
  Katinas}.} \bibinfo{year}{1977}\natexlab{}.
\newblock \showarticletitle{[Glossary of chronobiology (author's transl)]}.
\newblock \bibinfo{journal}{\emph{Chronobiologia}}  \bibinfo{volume}{4 Suppl 1}
  (\bibinfo{year}{1977}), \bibinfo{pages}{1—189}.
\newblock
\showISSN{0390-0037}


\bibitem[Hallac et~al\mbox{.}(2017)]%
        {hallac2017toeplitz}
\bibfield{author}{\bibinfo{person}{David Hallac}, \bibinfo{person}{Sagar Vare},
  \bibinfo{person}{Stephen Boyd}, {and} \bibinfo{person}{Jure Leskovec}.}
  \bibinfo{year}{2017}\natexlab{}.
\newblock \showarticletitle{Toeplitz inverse covariance-based clustering of
  multivariate time series data}. In \bibinfo{booktitle}{\emph{Proceedings of
  the 23rd ACM SIGKDD International Conference on Knowledge Discovery and Data
  Mining}}. \bibinfo{pages}{215--223}.
\newblock


\bibitem[Hawkes(1971)]%
        {hawkes_hawkes}
\bibfield{author}{\bibinfo{person}{A. Hawkes}.}
  \bibinfo{year}{1971}\natexlab{}.
\newblock \showarticletitle{Spectra of Some Self-Exciting and Mutually Exciting
  Point Processes}.
\newblock \bibinfo{journal}{\emph{Biometrika}}  \bibinfo{volume}{58}
  (\bibinfo{date}{April} \bibinfo{year}{1971}), \bibinfo{pages}{83}.
\newblock


\bibitem[Hira and Gillies(2015)]%
        {random_forest_feat_selection}
\bibfield{author}{\bibinfo{person}{Z. Hira} {and} \bibinfo{person}{D.F.
  Gillies}.} \bibinfo{year}{2015}\natexlab{}.
\newblock \showarticletitle{A Review of Feature Selection and Feature
  Extraction Methods Applied on Microarray Data}.
\newblock   \bibinfo{volume}{2015} (\bibinfo{date}{July} \bibinfo{year}{2015}),
  \bibinfo{pages}{1--13}.
\newblock


\bibitem[Inc.({[n.\,d.]})]%
        {fitbit}
\bibfield{author}{\bibinfo{person}{Fitbit Inc.}}
  \bibinfo{year}{[n.\,d.]}\natexlab{}.
\newblock \bibinfo{booktitle}{\emph{Fitbit Charge $2$}}.
\newblock
\urldef\tempurl%
\url{https://www.fitbit.com/us/charge2}
\showURL{%
\tempurl}


\bibitem[Jati et~al\mbox{.}(2021)]%
        {jati2021temporal}
\bibfield{author}{\bibinfo{person}{Arindam Jati}, \bibinfo{person}{Amrutha
  Nadarajan}, \bibinfo{person}{Raghuveer Peri}, \bibinfo{person}{Karel
  Mundnich}, \bibinfo{person}{Tiantian Feng}, \bibinfo{person}{Benjamin
  Girault}, {and} \bibinfo{person}{Shrikanth Narayanan}.}
  \bibinfo{year}{2021}\natexlab{}.
\newblock \showarticletitle{Temporal dynamics of workplace acoustic scenes:
  Egocentric analysis and prediction}.
\newblock \bibinfo{journal}{\emph{IEEE/ACM Transactions on Audio, Speech, and
  Language Processing}}  \bibinfo{volume}{29} (\bibinfo{year}{2021}),
  \bibinfo{pages}{756--769}.
\newblock


\bibitem[{K. Hui, and R. Sherratt}(2018)]%
        {emotion_hui}
\bibfield{author}{\bibinfo{person}{{K. Hui, and R. Sherratt}}.}
  \bibinfo{year}{2018}\natexlab{}.
\newblock \showarticletitle{Coverage of Emotion Recognition for Common Wearable
  Biosensors}.
\newblock \bibinfo{journal}{\emph{Biosensors}} \bibinfo{volume}{8},
  \bibinfo{number}{2} (\bibinfo{year}{2018}).
\newblock


\bibitem[Kasteren et~al\mbox{.}(2008)]%
        {home_setting}
\bibfield{author}{\bibinfo{person}{T.V. Kasteren}, \bibinfo{person}{A. Noulas},
  \bibinfo{person}{G. Englebienne}, \bibinfo{person}{}, {and}
  \bibinfo{person}{B. Kr\"{o}se}.} \bibinfo{year}{2008}\natexlab{}.
\newblock \showarticletitle{Accurate Activity Recognition in a Home Setting}.
  In \bibinfo{booktitle}{\emph{Proceedings of the 10th International Conference
  on Ubiquitous Computing}} (Seoul, Korea) \emph{(\bibinfo{series}{UbiComp
  '08})}. \bibinfo{publisher}{ACM}, \bibinfo{pages}{1--9}.
\newblock


\bibitem[Lawton and Brody(1969)]%
        {adl_lawton}
\bibfield{author}{\bibinfo{person}{M. Lawton} {and} \bibinfo{person}{E.M.
  Brody}.} \bibinfo{year}{1969}\natexlab{}.
\newblock \showarticletitle{{Assessment of Older People: Self-Maintaining and
  Instrumental Activities of Daily Living1}}.
\newblock \bibinfo{journal}{\emph{The Gerontologist}}  \bibinfo{volume}{9}
  (\bibinfo{date}{10} \bibinfo{year}{1969}), \bibinfo{pages}{179--186}.
\newblock


\bibitem[Michael et~al\mbox{.}(2006)]%
        {gps_michael}
\bibfield{author}{\bibinfo{person}{Katina Michael}, \bibinfo{person}{Andrew
  McNamee}, \bibinfo{person}{Michael~G Michael}, {and} \bibinfo{person}{Holly
  Tootell}.} \bibinfo{year}{2006}\natexlab{}.
\newblock \showarticletitle{Location-based intelligence-modeling behavior in
  humans using GPS}. In \bibinfo{booktitle}{\emph{2006 IEEE International
  Symposium on Technology and Society}}. IEEE, \bibinfo{pages}{1--8}.
\newblock


\bibitem[Mohammad and Nishida(2010)]%
        {wear_social_mohammad}
\bibfield{author}{\bibinfo{person}{Y. Mohammad} {and} \bibinfo{person}{T.
  Nishida}.} \bibinfo{year}{2010}\natexlab{}.
\newblock \showarticletitle{Using physiological signals to detect natural
  interactive behavior}.
\newblock \bibinfo{journal}{\emph{Applied Intelligence}} \bibinfo{volume}{33},
  \bibinfo{number}{1} (\bibinfo{date}{Aug.} \bibinfo{year}{2010}),
  \bibinfo{pages}{79--92}.
\newblock


\bibitem[Mundnich et~al\mbox{.}(2020)]%
        {mundnich2020tiles}
\bibfield{author}{\bibinfo{person}{Karel Mundnich}, \bibinfo{person}{Brandon~M
  Booth}, \bibinfo{person}{Michelle l’Hommedieu}, \bibinfo{person}{Tiantian
  Feng}, \bibinfo{person}{Benjamin Girault}, \bibinfo{person}{Justin
  L’hommedieu}, \bibinfo{person}{Mackenzie Wildman}, \bibinfo{person}{Sophia
  Skaaden}, \bibinfo{person}{Amrutha Nadarajan}, \bibinfo{person}{Jennifer~L
  Villatte}, {et~al\mbox{.}}} \bibinfo{year}{2020}\natexlab{}.
\newblock \showarticletitle{TILES-2018, a longitudinal physiologic and
  behavioral data set of hospital workers}.
\newblock \bibinfo{journal}{\emph{Scientific Data}} \bibinfo{volume}{7},
  \bibinfo{number}{1} (\bibinfo{year}{2020}), \bibinfo{pages}{354}.
\newblock


\bibitem[OMSignal({[n.\,d.]})]%
        {om_signal}
\bibfield{author}{\bibinfo{person}{OMSignal}.}
  \bibinfo{year}{[n.\,d.]}\natexlab{}.
\newblock \bibinfo{booktitle}{\emph{OMSignal}}.
\newblock
\urldef\tempurl%
\url{https://omsignal.com/}
\showURL{%
\tempurl}


\bibitem[Paparrizos and Gravano(2015)]%
        {paparrizos2015k}
\bibfield{author}{\bibinfo{person}{John Paparrizos} {and} \bibinfo{person}{Luis
  Gravano}.} \bibinfo{year}{2015}\natexlab{}.
\newblock \showarticletitle{k-shape: Efficient and accurate clustering of time
  series}. In \bibinfo{booktitle}{\emph{Proceedings of the 2015 ACM SIGMOD
  international conference on management of data}}.
  \bibinfo{pages}{1855--1870}.
\newblock


\bibitem[Paparrizos and Gravano(2017)]%
        {paparrizos2017fast}
\bibfield{author}{\bibinfo{person}{John Paparrizos} {and} \bibinfo{person}{Luis
  Gravano}.} \bibinfo{year}{2017}\natexlab{}.
\newblock \showarticletitle{Fast and accurate time-series clustering}.
\newblock \bibinfo{journal}{\emph{ACM Transactions on Database Systems (TODS)}}
  \bibinfo{volume}{42}, \bibinfo{number}{2} (\bibinfo{year}{2017}),
  \bibinfo{pages}{1--49}.
\newblock


\bibitem[Patel et~al\mbox{.}(2012)]%
        {patel_sensor_review}
\bibfield{author}{\bibinfo{person}{S. Patel}, \bibinfo{person}{H. Park},
  \bibinfo{person}{P. Bonato}, \bibinfo{person}{L. Chan}, {and}
  \bibinfo{person}{M. Rodgers}.} \bibinfo{year}{2012}\natexlab{}.
\newblock \showarticletitle{A review of wearable sensors and systems with
  application in rehabilitation}.
\newblock \bibinfo{journal}{\emph{Journal of NeuroEngineering and
  Rehabilitation}} \bibinfo{volume}{9}, \bibinfo{number}{1} (\bibinfo{date}{20
  Apr} \bibinfo{year}{2012}), \bibinfo{pages}{21}.
\newblock


\bibitem[Saeb et~al\mbox{.}(2015)]%
        {saeb2015mobile}
\bibfield{author}{\bibinfo{person}{Sohrab Saeb}, \bibinfo{person}{Mi Zhang},
  \bibinfo{person}{Christopher~J Karr}, \bibinfo{person}{Stephen~M Schueller},
  \bibinfo{person}{Marya~E Corden}, \bibinfo{person}{Konrad~P Kording},
  \bibinfo{person}{David~C Mohr}, {et~al\mbox{.}}}
  \bibinfo{year}{2015}\natexlab{}.
\newblock \showarticletitle{Mobile phone sensor correlates of depressive
  symptom severity in daily-life behavior: an exploratory study}.
\newblock \bibinfo{journal}{\emph{Journal of medical Internet research}}
  \bibinfo{volume}{17}, \bibinfo{number}{7} (\bibinfo{year}{2015}),
  \bibinfo{pages}{e4273}.
\newblock


\bibitem[Sano and Picard(2013)]%
        {stress_physio_sano}
\bibfield{author}{\bibinfo{person}{A. Sano} {and} \bibinfo{person}{R.~W.
  Picard}.} \bibinfo{year}{2013}\natexlab{}.
\newblock \showarticletitle{Stress Recognition Using Wearable Sensors and
  Mobile Phones}. In \bibinfo{booktitle}{\emph{2013 Humaine Association
  Conference on Affective Computing and Intelligent Interaction}}.
  \bibinfo{pages}{671--676}.
\newblock


\bibitem[Soto and John(2017)]%
        {big_five_soto}
\bibfield{author}{\bibinfo{person}{C. Soto} {and} \bibinfo{person}{O.P. John}.}
  \bibinfo{year}{2017}\natexlab{}.
\newblock \showarticletitle{The Next Big Five Inventory (BFI-2): Developing and
  Assessing a Hierarchical Model With 15 Facets to Enhance Bandwidth, Fidelity,
  and Predictive Power}.
\newblock \bibinfo{journal}{\emph{Journal of Personality and Social
  Psychology}}  \bibinfo{volume}{113} (\bibinfo{date}{July}
  \bibinfo{year}{2017}), \bibinfo{pages}{117--143}.
\newblock


\bibitem[Sutin et~al\mbox{.}(2016)]%
        {con_sed_sutin}
\bibfield{author}{\bibinfo{person}{A. Sutin}, \bibinfo{person}{Y. Stephan},
  \bibinfo{person}{M. Luchetti}, \bibinfo{person}{A. Artese},
  \bibinfo{person}{A. Oshio}, {and} \bibinfo{person}{A. Terracciano}.}
  \bibinfo{year}{2016}\natexlab{}.
\newblock \showarticletitle{The five-factor model of personality and physical
  inactivity: A meta-analysis of 16 samples}.
\newblock \bibinfo{journal}{\emph{Journal of Research in Personality}}
  \bibinfo{volume}{63} (\bibinfo{year}{2016}), \bibinfo{pages}{22 -- 28}.
\newblock


\bibitem[Viv{\'o}-Truyols and Schoenmakers(2006)]%
        {vivo2006automatic}
\bibfield{author}{\bibinfo{person}{Gabriel Viv{\'o}-Truyols} {and}
  \bibinfo{person}{Peter~J Schoenmakers}.} \bibinfo{year}{2006}\natexlab{}.
\newblock \showarticletitle{Automatic selection of optimal Savitzky- Golay
  smoothing}.
\newblock \bibinfo{journal}{\emph{Analytical chemistry}} \bibinfo{volume}{78},
  \bibinfo{number}{13} (\bibinfo{year}{2006}), \bibinfo{pages}{4598--4608}.
\newblock


\bibitem[Wadley et~al\mbox{.}(2008)]%
        {activity_wadley}
\bibfield{author}{\bibinfo{person}{V.G. Wadley}, \bibinfo{person}{O. Okonkwo},
  \bibinfo{person}{M. Crowe}, {and} \bibinfo{person}{L.A. Ross-Meadows}.}
  \bibinfo{year}{2008}\natexlab{}.
\newblock \showarticletitle{Mild Cognitive Impairment and Everyday Function:
  Evidence of Reduced Speed in Performing Instrumental Activities of Daily
  Living}.
\newblock \bibinfo{journal}{\emph{The American Journal of Geriatric
  Psychiatry}} \bibinfo{volume}{16}, \bibinfo{number}{5}
  (\bibinfo{year}{2008}), \bibinfo{pages}{416 -- 424}.
\newblock


\bibitem[Watson and Clark(1999)]%
        {panas_watson}
\bibfield{author}{\bibinfo{person}{D. Watson} {and} \bibinfo{person}{L.
  Clark}.} \bibinfo{year}{1999}\natexlab{}.
\newblock \showarticletitle{The PANAS-X: Manual for the positive and negative
  affect schedule-expanded form}.
\newblock  (\bibinfo{date}{January} \bibinfo{year}{1999}).
\newblock


\bibitem[Xu et~al\mbox{.}(2016)]%
        {hawkes_xu}
\bibfield{author}{\bibinfo{person}{Hongteng Xu}, \bibinfo{person}{Mehrdad
  Farajtabar}, {and} \bibinfo{person}{Hongyuan Zha}.}
  \bibinfo{year}{2016}\natexlab{}.
\newblock \showarticletitle{Learning Granger causality for Hawkes processes}.
  In \bibinfo{booktitle}{\emph{International conference on machine learning}}.
  PMLR, \bibinfo{pages}{1717--1726}.
\newblock


\bibitem[Yeh et~al\mbox{.}(2016)]%
        {yeh2016matrix}
\bibfield{author}{\bibinfo{person}{Chin-Chia~Michael Yeh}, \bibinfo{person}{Yan
  Zhu}, \bibinfo{person}{Liudmila Ulanova}, \bibinfo{person}{Nurjahan Begum},
  \bibinfo{person}{Yifei Ding}, \bibinfo{person}{Hoang~Anh Dau},
  \bibinfo{person}{Diego~Furtado Silva}, \bibinfo{person}{Abdullah Mueen},
  {and} \bibinfo{person}{Eamonn Keogh}.} \bibinfo{year}{2016}\natexlab{}.
\newblock \showarticletitle{Matrix profile I: all pairs similarity joins for
  time series: a unifying view that includes motifs, discords and shapelets}.
  In \bibinfo{booktitle}{\emph{2016 IEEE 16th international conference on data
  mining (ICDM)}}. Ieee, \bibinfo{pages}{1317--1322}.
\newblock


\bibitem[Yu and Sano(2023)]%
        {yu2023semi}
\bibfield{author}{\bibinfo{person}{Han Yu} {and} \bibinfo{person}{Akane Sano}.}
  \bibinfo{year}{2023}\natexlab{}.
\newblock \showarticletitle{Semi-Supervised Learning for Wearable-based
  Momentary Stress Detection in the Wild}.
\newblock \bibinfo{journal}{\emph{Proceedings of the ACM on Interactive,
  Mobile, Wearable and Ubiquitous Technologies}} \bibinfo{volume}{7},
  \bibinfo{number}{2} (\bibinfo{year}{2023}), \bibinfo{pages}{1--23}.
\newblock


\bibitem[Yue et~al\mbox{.}(2022)]%
        {yue2022ts2vec}
\bibfield{author}{\bibinfo{person}{Zhihan Yue}, \bibinfo{person}{Yujing Wang},
  \bibinfo{person}{Juanyong Duan}, \bibinfo{person}{Tianmeng Yang},
  \bibinfo{person}{Congrui Huang}, \bibinfo{person}{Yunhai Tong}, {and}
  \bibinfo{person}{Bixiong Xu}.} \bibinfo{year}{2022}\natexlab{}.
\newblock \showarticletitle{Ts2vec: Towards universal representation of time
  series}. In \bibinfo{booktitle}{\emph{Proceedings of the AAAI Conference on
  Artificial Intelligence}}, Vol.~\bibinfo{volume}{36}.
  \bibinfo{pages}{8980--8987}.
\newblock


\bibitem[Y{\"u}r{\"u}ten et~al\mbox{.}(2014)]%
        {routine_yürüten}
\bibfield{author}{\bibinfo{person}{Onur Y{\"u}r{\"u}ten},
  \bibinfo{person}{Jiyong Zhang}, {and} \bibinfo{person}{Pearl Pu}.}
  \bibinfo{year}{2014}\natexlab{}.
\newblock \showarticletitle{Decomposing activities of daily living to discover
  routine clusters}. In \bibinfo{booktitle}{\emph{Proceedings of the AAAI
  Conference on Artificial Intelligence}}, Vol.~\bibinfo{volume}{28}.
\newblock


\bibitem[Zhang et~al\mbox{.}(2023)]%
        {zhang2023self}
\bibfield{author}{\bibinfo{person}{Kexin Zhang}, \bibinfo{person}{Qingsong
  Wen}, \bibinfo{person}{Chaoli Zhang}, \bibinfo{person}{Rongyao Cai},
  \bibinfo{person}{Ming Jin}, \bibinfo{person}{Yong Liu},
  \bibinfo{person}{James Zhang}, \bibinfo{person}{Yuxuan Liang},
  \bibinfo{person}{Guansong Pang}, \bibinfo{person}{Dongjin Song},
  {et~al\mbox{.}}} \bibinfo{year}{2023}\natexlab{}.
\newblock \showarticletitle{Self-Supervised Learning for Time Series Analysis:
  Taxonomy, Progress, and Prospects}.
\newblock \bibinfo{journal}{\emph{arXiv preprint arXiv:2306.10125}}
  (\bibinfo{year}{2023}).
\newblock


\bibitem[Zhang and Chen(2014)]%
        {zhang_data_loss}
\bibfield{author}{\bibinfo{person}{Q. Zhang} {and} \bibinfo{person}{Z. Chen}.}
  \bibinfo{year}{2014}\natexlab{}.
\newblock \showarticletitle{A weighted kernel possibilistic c-means algorithm
  based on cloud computing for clustering big data}.
\newblock \bibinfo{journal}{\emph{International Journal of Communication
  Systems}} \bibinfo{volume}{27}, \bibinfo{number}{9} (\bibinfo{year}{2014}),
  \bibinfo{pages}{1378--1391}.
\newblock


\end{thebibliography}


\end{document}